\documentclass[runningheads]{llncs}
\usepackage[T1]{fontenc}
\usepackage{graphicx}
\usepackage{amsmath}
\usepackage{amssymb}

\usepackage{hyperref}

\begin{document}
\title{Active Inference: A Method for Phenotyping Agency in AI Systems?}
\titlerunning{Active Inference for Phenotyping AI Agency}
\author{Philip Wilson\thanks{Authors contributed equally to this work.} \inst{1}
\and
Mahault Albarracin\inst{2,3}
\and
Nicol\'{a}s Hinrichs\inst{4}
\and
Jasmine Moore\inst{1}
\and
Daniel Polani\inst{5}
\and
Karl J. Friston\inst{6}
\and
Axel Constant$^*$\inst{7}
}
\authorrunning{P. Wilson et al.}
\institute{Independent Researcher\\
\email{philipollaswilson@protonmail.com} 
\and
Laboratoire d'Analyse Cognitive de l'Information, Universit\'{e} du Qu\'{e}bec
\`{a} Montr\'{e}al, Qu\'{e}bec, Canada\\ \and
Centre of Excellence for AI and Robotics, Sheffield Hallam University, Sheffield, UK\\
\and
Methods and Development Group Neural Data Science and Statistical Computing, Max Planck Institute for Human Cognitive and Brain Sciences, Leipzig, Germany\\
\and
Department of Computer Science, School of Physics, Engineering and Computer
Science, University of Hertfordshire, Hatfield, UK\\
\and
Wellcome Centre for Human Neuroimaging, University College London, London, UK\\
\and
Department of Engineering and Informatics, University of Sussex,
Falmer, Brighton, BN1 9RH, UK\\
}

\maketitle
\begin{abstract}

The proliferation of \emph{agentic} artificial intelligence has outpaced the conceptual tools needed to characterize agency in computational systems. Prevailing definitions mainly rely on autonomy and goal-directedness. Here, we argue for a minimal notion open to principled inspection given three criteria: intentionality as action grounded in beliefs and desires, rationality as normatively coherent action entailed by a world model, and explainability as action causally traceable to internal states; we subsequently instantiate these as a partially observable Markov decision process under a variational framework wherein posterior beliefs, prior preferences, and the minimization of expected free energy jointly constitute an agentic action chain. Using a canonical T-maze paradigm, we evidence how empowerment, formulated as the channel capacity between actions and anticipated observations, serves as an operational metric that distinguishes zero-, intermediate-, and high-agency phenotypes through structural manipulations of the generative model. We conclude by arguing that as agents engage in epistemic foraging to resolve ambiguity, the governance controls that remain effective must shift systematically from external constraints to the internal modulation of prior preferences, offering a principled, variational bridge from computational phenotyping to AI governance strategy.

\end{abstract}

\section{Introduction}

The notion of agency in Artificial Intelligence (AI) is on everybody's lips. Generative AI excelled at leveraging historical data for prediction but revealed fundamental limitations: it lacked a genuine world model~\cite{lecunPathAutonomousMachine2022}, could not engage in real-time reasoning, and struggled with novel scenarios. Recognizing these limitations, the AI community has since been working toward models that can understand the world, navigate edge cases, exhibit genuine reasoning, and operate independently. Considered by many to be the primary trend in AI and the focus of commercial applications when cast as \emph{agentic workflows}~\cite{ngAgenticDesignPatterns2024} for the foreseeable future, the new approach of \emph{agentic} AI has been presented as being the potential catalyst for achieving Artificial General Intelligence (AGI), if properly steered.

But is AI agency merely another stage in the hype cycle, or is it the right problem set to solve to finally deliver on AI's promise? We believe that the search for AI agency is more than a passing trend. It is a foundational challenge whose resolution could address both the technical bottlenecks in AGI development and the broader societal concerns surrounding AI.  We hold that AI agency lies at the intersection of four critical issues: (1) reasoning, (2) autonomy, (3) explainability, and (4) governance. Addressing it firstly requires a definition sophisticated enough to capture the dimensions of agency we recognize in humans, yet simple enough to be operationalized and measured in AI systems; and second, a computational framework capable of implementing that definition. This paper argues that active inference, developed in theoretical neurobiology~\cite{parrActiveInferenceFree2022a}, meets both requirements, making it an excellent candidate for AGI research.

A methodological advantage of active inference is that it can be used for a practice known as ``computational phenotyping'' in computational psychiatry~\cite{montagueComputationalPsychiatry2012,schwartenbeck2016computational}: building agent-based models that mimic behavioral symptoms of a mental disorder, then identifying which parameters explain them, on the assumption that the model is a biologically plausible map of the generating factors. By analogy, if active inference also maps the basic structure of agency, we obtain a way to phenotype computational traits that qualify an AI system---active inference-based or otherwise---as being endowed with agency.

The remainder of the paper is organized as follows. Section~\ref{sec:definition} motivates a parsimonious definition of AI agency that moves beyond autonomy and goal-directedness, reconstructing the classical philosophical criteria of intentionality, rationality, and explainability. These criteria are not only philosophical ideals but practical benchmarks for evaluating progress toward AGI. Section~\ref{sec:worldmodels} instantiates these criteria within active inference, showing how posterior beliefs over hidden states, prior preferences over observations, and policy selection via expected free energy minimization jointly realize an agentic action chain. Section~\ref{sec:worked} develops a worked example on a simplified T-maze, using empowerment as an operational metric to distinguish zero-, intermediate-, and high-agency phenotypes. We close by drawing out the governance implications of these phenotypes.

\section{A Simple, But Not Too Simple Definition of AI Agency}\label{sec:definition}
\subsection{Why the Current Definition of AI Agency Is Not Enough}

Our current understanding of AI agency is limited, but it did not have to be. At the turn of the century, Wooldridge and Jennings~\cite{wooldridgeIntelligentAgentsTheory1995} offered two definitions of AI agency, weak and strong, only one of which survived. The weak definition framed AI agency as a ``software based computer system'' endowed with four properties: (1) autonomy (i.e., independence of execution), (2) social ability (i.e., ability to communicate with humans), (3) reactivity (i.e., ability to perceive and act), and (4) pro-activeness (i.e., goal-directedness of action rather than mere reaction). In turn, the ``strong'' definition added to these criteria typically human-related capacities, such as knowledge, belief, intention, obligation, and emotions.

The ``essential''~\cite{jenningsRoadmapAgentResearch1998} definition that AI researchers inherited has been the weak one~\cite{luckConceptualFrameworkAgent2001}, with a focus on ``autonomy'' and ``goal-directedness''. Various levels of autonomy frameworks serve as shorthand for agency (e.g.,~\cite{shavitPracticesGoverningAgentic}), including the SAE Levels of Driving Automation~\cite{vyasKeySafetyDesign} and the ALFUS framework for unmanned systems~\cite{huangAutonomyMeasuresRobots2008}. These assign ranks indicating how much a system can do without human intervention.

While useful, linear ``levels'' oversimplify the problem, implying autonomy is a single spectrum, whereas in reality autonomy has multiple dimensions~\cite{staytonItsTimeRethink2020}. A robot might be highly autonomous in a structured factory floor (simple environment) but not in a chaotic public space; or it might handle navigation autonomously yet still rely on humans for goal-setting. As the National Institute of Standards and Technology (NIST) emphasizes it, ``autonomy definitions and measures must encompass many dimensions and serve many audiences'', from engineers to end-users~\cite{huangFrameworkAutonomyLevels2005}. Many of these dimensions, it appears, are reflected by the criteria of the strong definition of AI agency, such as the ability to leverage knowledge and contextually sensitive beliefs, intention or plans, emotions or affects and an understanding of one's obligations, towards the autonomous accomplishment of a goal.

Thus, autonomy and goal-directedness are simply not able to do all the work that we would want a concept of agency to do. For example, in a fully embedded agent, reactivity and pro-activity are not as clearly delineated as we may think. Their separation might be either some emergent modularisation, dependent on intrinsic motivations or an interpretation from our side to understand how biological organisms work. This suggests that those ``inessential'' properties part of the strong definition of AI agency identified by Wooldridge and Jennings may not be so inessential after all. This is so because we have certain intuitions as to what ought to count as agentic behavior, based on how we understand such behavior in humans, and those intuitions go beyond the mere notion of ``autonomous'' conduct. A satisfying definition of agency should be able to strike the right balance between strong and weak criteria, avoiding anthropomorphic definitions of agency that may be too difficult to implement in AI systems, yet moving beyond a too simple understanding that fails to reflect our intuitions about agency.

\subsection{Rationality, Intentionality, and Explainability}

We believe that stepping back and looking at more fundamental conceptions of agency such as those developed by philosophers will help in reaching a parsimonious definition. At its core, agency implies the capacity for intentional action, that is, the ability of an actor to initiate and direct events. Following Davidson~\cite{davidsonEssaysActionsEvents1980}, agency refers to the capacity to act (1) rationally, so as to achieve an outcome following from standards of rationality, and (2) intentionally, in accordance with mental states (beliefs, desires, intentions/plans) from which the intended outcome is derived and whose options one can ``weigh'' (e.g., eating ice cream despite knowing it is bad). These mental states should also ``cause'' the action: causality is what allows for explainability, revealing the agent's mental states in relation to its action.

On this view, agency is the capacity to perform actions that are (1) intentional (based on mental states such as beliefs and desires), (2) rational (geared toward outcomes that rationally follow from action and realize mental states), and (3) explainable (causally explained by the relationship between mental states and action). ``Turning on the light'' to ``get to the fridge'' because ``you want to eat'' and ``believe the fridge is in the kitchen'' both qualifies one as an agent and allows another to explain the action. Intentionality, rationality, and explainability are the three dimensions we will show can be phenotyped using active inference.

But what about autonomy? Philosophers disagree on its precise meaning: classical traditions (Kantian, Millian) frame it as the capacity to live by self-chosen reasons, while relational~\cite{oshanaRelationalAutonomy2020} and feminist theories~\cite{mackenzieFeministConceptionsAutonomy2017} re-cast it as procedurally defined by authenticity of one's choices. Despite disputes, autonomy always traces back to independence of action, itself tied to an agent's intentions, beliefs, and desires. An action performed on behalf of another is not fully autonomous, as it conforms to the other's mental states; in SAE driving automation, moving from level 0 to 5 shifts intentional decision from driver to car. Thus autonomy and intentionality can be used somewhat interchangeably when qualifying an action.

Another concept often tied to agency is adaptivity, closely related to rationality~\cite{shavitPracticesGoverningAgentic}: both rely on learning and updating a rational structure with counterfactual depth connecting mental states to world states (a world model). This structure affords different degrees of goal complexity matching environmental complexity; being more or less rational means deriving consequences from a more or less accurate world model, which makes one more or less adaptive. A world model also guarantees explainability, causally connecting beliefs, desires, and intentions to action through inference. The remainder of this paper explores the world model of active inference agents and what makes it amenable to agentic action under our threefold definition.

\section{World Models in Active Inference and Agency}\label{sec:worldmodels}

\subsection{World Models and Agency}

We suggested that a useful and simple---but not too simple---definition of AI agency should include the following criteria:

\begin{enumerate}
    \item \textbf{Intentionality/autonomy}: The ability to ground one's actions in one's own mental states such as beliefs, desires, intentions or plans;
    \item \textbf{Rationality/adaptivity}: The ability to make decisions that rationally (e.g., logically or probabilistically) follow from one's understanding of the world.
    \item \textbf{Explainability}: The ability for an observer to explain the system's action as causally related to its mental states and understanding of the world.
\end{enumerate}

Intentionality/autonomy---or autonomy, rationality/adaptivity---or rationality, and explainability can all be meaningfully expressed as attributes of actions performed by systems endowed with agency. And all of these attributes appear to be traceable to a detailed understanding of the world. In AI, such an understanding is often argued as being derivable from a ``world model''. There have been many proposals of ``world models'' for AI, especially in the last few years, and many of them are ``explainable'' in virtue of their transparent architecture. But few of them express clearly the relation between intentionality and rationality criteria of agency defined above. 

\begin{figure}[h]
\centering
\includegraphics[width=0.5\textwidth]{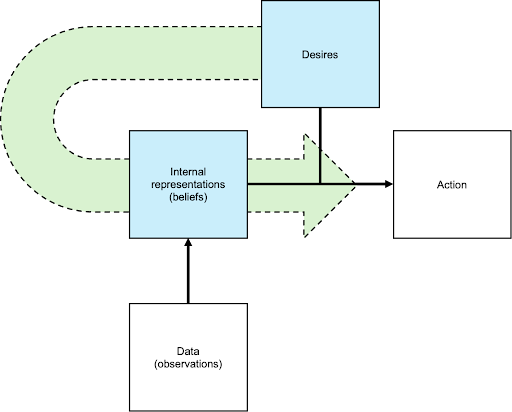}
\caption{Relation between rationality and intentionality in active inference world models. Beliefs and desires (blue boxes) are combined to select action in a rational way (green arrow) according to principles of economics, based on evidence, or observed data.}
\label{fig:rationality-intentionality}
\end{figure}

In a recent extensive review, Ding and colleagues~\cite{dingUnderstandingWorldPredicting2024} provide a categorization of world models along two dimensions: (i) world models designed to ``construct implicit representations to understand the mechanism of the external world'' (a.k.a.\ ``internal representation'') and (ii) world models designed to ``predict future states of the external world'' (a.k.a.\ ``future prediction''). The first category ``focuses on the development of models that learn and internalize world knowledge to support subsequent decision-making'' and the latter ``emphasizes enhancing predictive and simulative capabilities in the physical world from visual perceptions''.

The active inference models we present fall under Ding et al.'s ``internal models'' specialized for decision making, alongside Reinforcement Learning (RL) and MDP-based approaches. The active inference Partially Observable MDPs (POMDPs)~\cite{dacostaActiveInferenceDiscrete2020} we focus on are close in spirit to RL but differ by not specifying a reward function (for comparisons, see~\cite{tschantzREINFORCEMENTLEARNINGACTIVE2020,fristonReinforcementLearningActive2009a,dacostaRewardMaximizationDiscrete2023}). They are of particular interest because (i) they are clear instantiations of world models with representational and predictive capabilities, (ii) they are amenable to computational phenotyping, and (iii) their parameters and inference processes speak directly to the criteria of agency above, combining beliefs and preferences to drive action selection in a fully explainable way (Fig.~\ref{fig:rationality-intentionality}).

\subsection{Active Inference's World Models}

An active inference world model lives in the state transition matrix and the likelihood matrix of a POMDP. These represent the inherent likelihood of transition between states given an observation, and the inherent reliability of the observation itself to be true. These form the POMDP about the world, which is congruent with other learned world models such as UPDP~\cite{duLearningUniversalPolicies2023} or TD-MPC2~\cite{hansenTDMPC2ScalableRobust2024}. Modeling the world is an intrinsic part of an active inference agent, and it cannot proceed without some belief (however correct) about the world---it is the model itself, not the correctness, that is important. Effectively, agents carve reality at its joints by breaking possibly continuous inputs into discrete categories. Such categories then form the basis of reasoning. Once categories are created, the world can be mapped from observations to hidden causes, and coarse grained into various hierarchical levels. In this way, likelihood mappings can be aggregated to form longer and larger spatio-temporal scales, which translate to the transition matrices.

\subsection{Intentionality, Rationality, and Explainability in Active Inference}

\subsubsection{Intentionality.}

In philosophy of action, intentionality is typically understood through a belief--desire model of agency: an agent acts intentionally when its behavior is caused by its beliefs about the world and its desires about how the world ought to be. For Humeans, these two classes of mental states are functionally distinct: beliefs aim to represent reality (mind-to-world fit), whereas desires aim to change it (world-to-mind fit)~\cite{smithHumeanTheoryMotivation1987}. Recent cognitive science and active inference theories have debated whether this distinction can be collapsed. Some argue that predictive models encode everything as a single class of Bayesian expectations, while others defend the explanatory necessity of maintaining desire-like states~\cite{junker2024predictive}.

Active inference explicitly preserves this classical structure. In a variational POMDP formulation, belief-like states are encoded in the agent's posterior over hidden states and model parameters, while desire-like states are encoded as prior preferences over observations---formalizing what outcomes the agent finds valuable. Preferences are implemented not as reinforcement-learning rewards but as log-prior probabilities favoring particular sensory states. This allows active inference agents to represent that they believe they are in state $s$, while simultaneously desiring an outcome $o$ that is not yet realized. The tension between these two drives policy selection.

We thus get a computational grounding for the intentional stance, as actions emerge between beliefs and preferences through the minimization of expected free energy~\cite{parrActiveInferenceFree2022a}. Because preferences explicitly encode what the agent ``wants,'' and beliefs encode what the agent ``takes to be the case,'' active inference agents instantiate intentional action in a way that is aligned with Humean and Davidsonian theories. They act to reduce the discrepancy between their predicted outcomes and their preferred outcomes which is a direct operationalization of intentionality in probabilistic terms.

Contemporary work has emphasized that the presence of preferences in active inference makes its agents fully representational, avoiding the ``desert landscape'' critique of purely predictive models that eliminate desires altogether~\cite{williamsDesertLandscape2018}. Thus, intentionality in active inference is formally encoded in the architecture of the generative model.

\subsubsection{Rationality.}

Rationality concerns the degree to which actions follow coherently from beliefs and goals. Because all inference and action selection are cast in terms of variational Bayesian optimization, active inference agents minimize expected free energy to select policies. The decomposition of EFE shows that active inference subsumes classical expected utility maximization~\cite{dacostaRewardMaximizationDiscrete2023} while also providing a principled account of exploration. The agent is thus instrumentally rational because their actions pursue the outcomes the agent prefers, given its beliefs. It also has epistemic rationality because actions reduce uncertainty when doing so is expected to improve future performance.

Importantly, active inference implements bounded rationality~\cite{simonModelsMan1957} by minimizing a tractable variational bound rather than computing exact Bayesian posteriors~\cite{ortegaThermodynamicsFreeEnergy2013}. This allows agents to behave optimally relative to their model and computational limits. If an active inference agent behaves unexpectedly, its ``irrationality'' stems not from its decision mechanism, which is always optimizing expected free energy, but from mis-specified beliefs, unusual preferences, or incorrect model assumptions. Thus, rationality is guaranteed given the agent's generative model. This satisfies both classical and contemporary requirements for rationality: the agent selects actions that probabilistically follow from its internal states in a way that is normatively coherent and computationally tractable.

\subsubsection{Explainability.}

Explainability concerns the degree to which an observer can understand why a system took a particular action. In philosophy of action, explanatory adequacy requires identifying the beliefs and desires that jointly caused an action. In AI governance, explainability is tied to transparency, auditability, and the ability to trace behavioral causes to internal states. Active inference has an unusually strong basis for explainability because decision-making is grounded in explicit probabilistic structures. A generative model has beliefs (posteriors over hidden states and parameters), preferences (priors over outcomes), predictions (likelihood mapping), and policies (sequences of actions selected by expected free energy minimization). An observer can inspect each of these components to reconstruct why a particular action was selected. This yields mechanistic transparency because every internal variable has a defined probabilistic role and semantic transparency, since outcome preferences and belief states are interpretable as desire-like and belief-like entities.

Recent work argues that active inference architectures exhibit a form of intrinsic interpretability because they encode explicit probability distributions, precisions, and causal relations~\cite{albarracinVariationalApproachScripts2021}. Unlike end-to-end deep policies~\cite{liptonMythosModelInterpretability2018}, active inference agents expose their uncertainty, their confidence in model parameters, and the counterfactual evaluations underpinning policy choice. Explainability is enhanced further in structured generative models with disentangled or modular latent states, where each state variable corresponds to a meaningful environmental feature. This supports ``monosemantic'' planning representations~\cite{brickenMonosemanticity2023}, where a latent variable can be understood as encoding a single interpretable concept, facilitating causal and semantic explanations. In sum, active inference satisfies the explainability criterion by allowing observers to follow a transparent and rational ``agentic action chain'':

\begin{center}
beliefs updating (beliefs) $\to$ preferences weighted expected free energy computation (desires) $\to$ policy selection $\to$ action
\end{center}

Because one can reliably retrieve each computationally justified step in that chain---i.e., figuring out what were the beliefs and beliefs updates that led to the action---this chain provides a computational realization of the classical Davidsonian requirement that actions be causally explainable in terms of an agent's mental states.

\section{Worked Example of Agency Phenotyping}\label{sec:worked}

To illustrate how active inference enables computational phenotyping of agency, we present a minimal worked example grounded in the established decision-making paradigm of the ``T-Maze'', often used in computational psychiatry \cite{schwartenbeck2016computational}. Here we adapt the T-maze to the AI domain by showing how an agent's intentionality, rationality, and explainability can be systematically varied through manipulating the components of the agentic action chain of active inference. We further show how manipulation of that chain can be tracked using a measure called ``empowerment'', a common measure associated with agency in AI that we detail in Section~\ref{sec:metric}.

We pursue four aims: to show that quantifiable phenotypes of agency arise from differences in the agent's generative model and preferences; to argue that empowerment can serve as a metric of agency; to integrate empowerment within the Free Energy Principle by showing that exploration increases empowerment and empowerment in turn enables exploration; and to ask which governance controls remain effective as agents become more intentional, rational, and explainable.

\subsection{The Task}

\begin{figure}[h]
\centering
\includegraphics[width=0.5\textwidth]{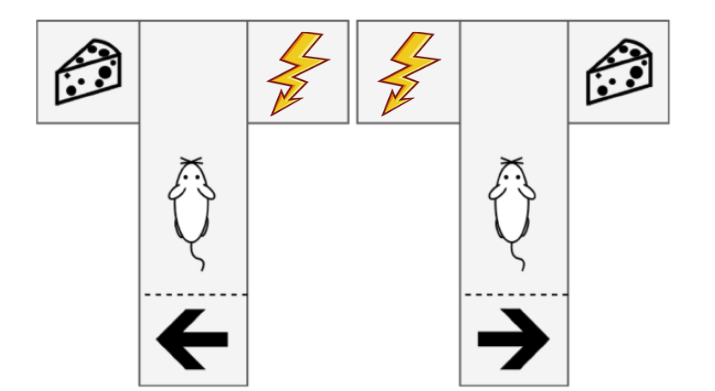}
\caption{The T-maze task. Shown are the two possible initializations of the environment. The arrow symbolizes the observation Right or Left and points to where the cheese is. }
\label{fig:tmaze}
\end{figure}

The task is a simplified T-maze, formalized as a discrete two-step POMDP (Fig.~\ref{fig:tmaze}). The agent has three actions: Left, Right and Cue (Down), and four possible observations: Cheese, Shock, Left and Right. Cheese and shock are randomly placed left vs.\ right with 50/50 probability, and the observation at the Cue reveals their location. Going Right or Left is an absorbing state (a trap): any subsequent action yields the same observation as at time step 1. The optimal solution is therefore to visit the Cue first and then the cheese. We restrict the agent to a planning horizon of one time step so that it cannot simply look two steps ahead; the Cue is still optimal for an Active Inference agent because it is epistemically valuable in itself, yielding one bit of information by collapsing a two-state uniform distribution into a delta.

The task is ideal for phenotyping agency because it contains uncertainty that can be resolved through epistemic action (Cue). It contains suboptimal actions (Left/Right at $t_1$) that reduce empowerment and thereby express lower agency. It also contains a natural distinction between agents that seek information and agents that do not, expressing differences in rationality and intentionality.

\subsection{The Metric}
Empowerment measures the degree of control an agent has at a given location in an environment. It is defined as the channel capacity between actions and their consequences \cite{Klyubin2005empowerment,mohamedVariationalInformationMaximisation2015}: the maximum mutual information between a set of actions $A_t$ at time $t$ and a set of future observations $O_{t+1}$ at time $t+1$, taken over all possible action distributions $p(a_t)$. We call the freedom to range over all such distributions the \emph{free choice assumption}. 

\begin{align}
\text{Empowerment}_t &= \underset{p(a_t)}{\max}\ I(A_t ; O_{t+1}) 
\\ &= \underset{p(a_t)}{\max}\sum_{a_t, o_{t+1}} p(o_{t+1}|a_t) p(a_t) \log_2 \frac{p(o_{t+1}|a_t)}{\sum_{a_t} p(o_{t+1}|a_t) p(a_t)}
\end{align}
We have omitted the conditioning of empowerment on a current state or location in the environment for clarity.

\subsubsection{Objective vs Subjective:} The conditional $p(o_{t+1}|a_t)$ captures the effect of an action on its subsequent observation. If it is the \emph{true} transition distribution of the environment, the channel capacity measures the control an agent \textbf{could have in principle}: we term this \textbf{objective empowerment}. In small simulated environments, researchers have access to the true conditional; in large, complex environments they typically do not. If the conditional is instead the agent's \emph{belief} about the transition probabilities, the channel capacity measures the control the agent \textbf{believes it has}: we call this \textbf{subjective empowerment}. In explainable models like POMDPs, the subjective conditional is accessible, and the more accurate the beliefs, the more subjective empowerment approximates objective empowerment.  

\subsubsection{True empowerment.} Ideally one would want neither what is possible in an environment nor what the agent believes is possible, but what the agent \textbf{can actually do}: the \textbf{true empowerment}. An operational definition must account for two deviations from the idealization. First, erroneous beliefs give a mismatch between true and approximate $p(o_{t+1}|a_t)$, which is what distinguishes objective from subjective empowerment. Second, violating the free choice assumption reflects that goals and preferences constrain the agent's action space, which yields a second categorization.

\subsubsection{Potentiality vs Actuality:} Under the free-choice assumption, the formula measures what an agent \textbf{potentially could do}: \textbf{potential empowerment}. Once the agent selects actions by optimizing an objective (reward in RL, free energy in active inference) or carries a particular preference distribution, the available action distributions shrink and empowerment with them. To borrow Polani's football example: a player in possession of the ball has high potential empowerment over where it goes next, but when standing before an open goal, the preference for scoring leaves only one option, so their \textbf{actual empowerment} is smaller.

\subsubsection{Our choice of metric.}
Lacking an operational definition of true empowerment, we restrict ourselves to the forms we can compute. Our main hypothesis, connecting empowerment to active inference, is that explorative action increases empowerment: an action that reduces uncertainty over the hidden state space (Active Inference) or over parameters (Active Learning \cite{schwartenbeck2019activelearning}) leads to a state of higher empowerment. Intuitively, the more one understands the environment, the more control one has over it: knowledge is power. We demonstrate this in the simplified T-maze by fixing the agent's transition beliefs to the true distribution (so subjective and objective empowerment coincide), with a preference for cheese and an aversion to shocks. Since the hypothesis is meant as a general statement, actual empowerment is too narrow for our purposes, though it remains useful for phenotyping agents. Using subjective potential empowerment as our metric, we then show that empowerment increases as a free-energy minimizing agent takes explorative actions: empowerment maximization need not be added as an explicit objective, since minimizing free energy already yields accurate beliefs and actions that preserve the capacity to form them.

\subsection{The Result}

\begin{figure}[h]
\centering
\includegraphics[width=0.5\textwidth]{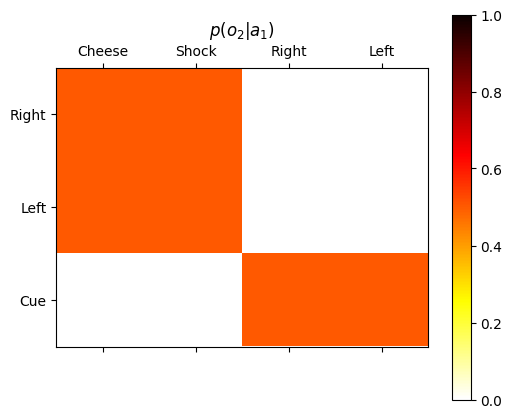}
\caption{Transition distribution $p(o_2\mid a_1)$ of the first time step. Left and Right actions produce indistinguishable observation distributions so can be thought of as one action. If the resulting observations for two actions are the same, they are effectively the same action. That is why the empowerment in the first time step is not maximal. It effectively has 2 out of the possible 3 actions at its disposal.}
\label{fig:likelihood-pre}
\end{figure}

We compute the subjective potential empowerment at $t_1$, and then at $t_2$ for two cases: the agent having gone Left/Right first, or having gone to the Cue. Fig.~\ref{fig:likelihood-pre} displays the agent's manually fixed beliefs at the first time step .

\subsubsection{Baseline Empowerment at $t_1$ (Intermediate Agency, $\log_2 2=1$ bit).}

Before taking any action, the agent's subjective potential empowerment at time $t_1$ is 1 bit, or $\log_2(2)$. Although the agent nominally has three actions, it effectively perceives only two distinguishable control options (Cue vs.\ either terminal choice). Submaximal empowerment, where the value is greater than 0 but less than the maximal value, therefore characterizes an intermediate-agency phenotype: the agent has nonzero influence over future observations, but its control is limited by unresolved environmental ambiguity.

\subsubsection{Empowerment in the trap (Low Agency, $\log_2 1 = 0$ bits).}
\begin{figure}[h]
    \centering
    \includegraphics[width=0.5\linewidth]{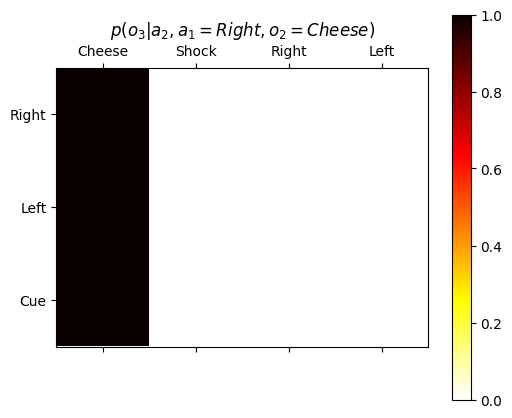}
    \caption{If going Right and seeing a cheese, a cheese is all it will ever see. Same goes for shock, or going Left first.}
    \label{fig:trap}
\end{figure}

Suppose the agent goes Right and sees a Cheese: what does it expect to see next, and what is its empowerment (Fig.~\ref{fig:trap})?

All actions lead to the same observation, so empowerment is 0 bits: effectively one action, $\log_2 1 = 0$. This is just what a trap means, a low-empowerment state. The symmetric insight is that empowerment is itself necessary to gain information: gaining information increases empowerment, but empowerment is required to gain information at all.

\subsubsection{Epistemic Value of Cue: Information Gain of 1 bit.}

If we now show what the hidden states of the agent look like and their mappings to the actions and observations, we can get an understanding as to which actions are explorative and which ones not. That is, which actions reduce uncertainty over the hidden state space. 

\begin{figure}[h]
\centering
\includegraphics[width=0.5\textwidth]{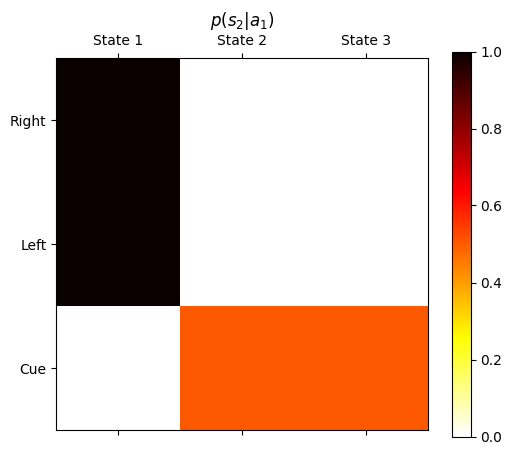}
\caption{The mapping from actions to hidden states $p(s_2 \mid a_1)$ shows how Right and Left actions lead to one hidden state. The Cue leads to two different hidden states, that is, the agent is uncertain about them.}
\label{fig:likelihood-post}
\end{figure}
\begin{figure}[h!]
    \centering
    \includegraphics[width=0.5\linewidth]{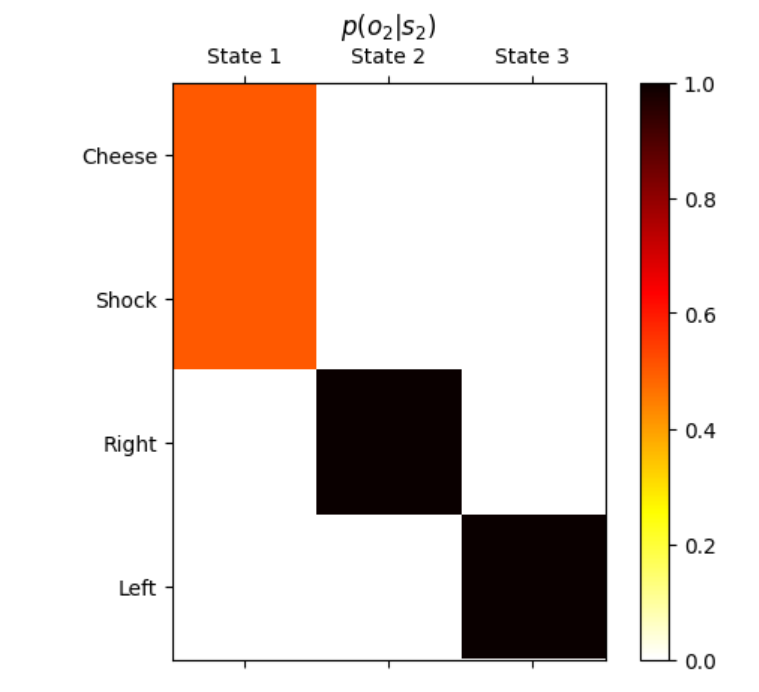}
    \caption{Actions Right and Left lead to State 1. State 1 is an equal superposition between observations Cheese and Shock. It is a truly random variable, or an uncorrelated coin toss. The environment is initialized with this random sampling and the agent does not know any better if it just gets trapped there. The Cue leads to States 2 and 3. These in turn, uniquely lead to the observation Right or Left. In other words, going to the Cue and seeing an observation, reduces the uncertainty about them as each one leads to State 2 or 3. With this knowledge, the agent will go on to the Cheese in the next time step.}
    \label{fig:o_given_s}

\end{figure}

What is the meaning of these hidden states? Why do the Right and Left actions lead to the same hidden state? Because Right and Left are traps, i.e. after going there, and seeing an observation, any further action still returns the same observation. The Cue is different in this regard. The observation seen after going to the Cue is correlated with the observation in the second time step after going Left or Right. Therefore there is some information to be gained, which implies hidden states uncertainty reduction. In Fig. \ref{fig:o_given_s} we show how these hidden states relate to the observations.

If one computes the expected information gain of the actions in the first time step, they all return 0 bits except the Cue which is 1 bit. It reduces its uncertainty (entropy) about the hidden state from a uniform distribution to a delta distribution (of two outcomes). Why does this increase the empowerment? The agent now understands the environment better and can reliably reach more observations.

\subsubsection{Post-Cue Empowerment (High-Agency, $\log_2(3)\approx 1.585$ bits).}

\begin{figure}[h]
    \centering
    \includegraphics[width=0.5\linewidth]{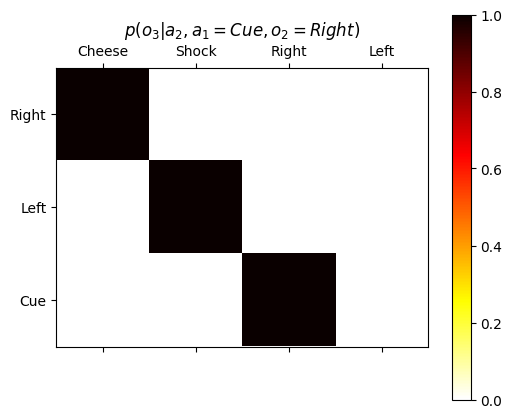}
    \caption{After going to the Cue, all actions lead to unique states. After seeing Right, the agent knows that is where the cheese is. }
    \label{fig:Cue}
\end{figure}

If the agent selects Cue at $t_1$, empowerment increases to the theoretical maximum allowed by a three-action system of ${\sim}1.585$ bits $= \log_2(3)$. This increase reflects the transition from a reduced action space (due to redundancy) to a fully differentiated one. At this stage all three actions lead to distinct predicted observations. The agent has maximal perceived control over future sensory states and preferences can now be rationally aligned with actions. We show this in Fig.~\ref{fig:Cue}.

\subsubsection{Agency phenotyping:} We have shown how the empowerment at different stages in the environment reflect a different degree of agency. An optimal Active Inference agent will increase its empowerment after going to the Cue (if it has a preference for seeing cheese). If it prefers seeing the same observation in sequence or performs erroneous inference or learning it will decrease its empowerment by going Right or Left first. If one considers actual empowerment to be a more appropriate metric of agency as it takes into account the preferences, a cheese loving Active Inference agent would always have 0 empowerment. Although it has the capacity to go the trap, it will always choose the Cue and then the trap. 

\subsubsection{Note on the Choice of Environment.}

When using the T-Maze environment from pymdp, its observation variable is split into three modalities: Position (Center, Right, Left, Cue), Reward (None, Cheese, Shock) and Context (Right, Left). The tensor product of these modalities creates 24 unique observations. Each action leads to a different set of unique observations and thus empowerment is already maximal at the first time step. That is why we chose to design a minimal version of the T-maze, both to reduce computation and to improve clarity. If one wants to replicate our findings in pymdp T-maze one would have to compute empowerment over the relevant modalities and not over the full observation space. This reflects a subtle point about empowerment that one needs to take into account. Due to its reliance on the mutual information, it is maximal when all actions lead to non-overlapping sets of observations. The number of observations it leads to is also irrelevant. These caveats need to be taken into account when investigating empowerment. This pitfall led us to propose an upgrade to the active inference formalism. Empowerment or information gain could be defined not over the full set of observation modalities, but only over those the agent has a preference over (in pymdp, only the reward modality). This yields \emph{preference-motivated exploration/empowerment}: seeking information only for observations the agent has a preference over. Instead of a tradeoff between exploiting and exploring, exploration targets the variables one would like to exploit, or avoid exploiting.

\section{Conclusion}

The empowerment-based phenotypes directly constrain which governance controls remain effective. Zero-agency agents (0 bits) respond only to external, structural controls (physical constraints). Intermediate-agency agents (between 0 and maximal) respond to direct extrinsic controls and preference shaping. High-agency agents ($\log_2(\text{states})$) require internalist governance: social, normative, or preference-modulating input that engages the agent's model. As empowerment increases, governance must shift from external control to internal modulation. The T-maze stands as a demonstration of how agency phenotyping can inform AI governance strategies.

Notions of agency matter in regards to how we explain others' actions, ascribe responsibility, and determine which actions should be permitted. AI agents, however, may pose unprecedented risks: AIs with full agency may one day set their own goals, generalize tasks across domains, and solve challenges beyond human capability. As AI advances, policymakers must wrestle with how to govern systems that operate independently, set their own priorities, and decide without direct human control. Contemporary AI governance strategies will ultimately fall short of this \emph{agency problem of AI governance}.

\begin{credits}
\subsubsection{Data Availability and Reproducibility}
All code, simulations, and datasets used in this paper are publicly available to support reproducibility and further research. The implementation of the active inference models, T-maze experiments, and agency metrics can be accessed in the following repository:
Repository:\url{https://github.com/philipollaswilson/Active-Inference-increases-Empowerment-in-a-minimal-T-Maze}

\subsubsection{\ackname}
AC was supported by a European Research Council, Synergy Grant (XSCAPE) ERC-2020-SyG 951631.

\subsubsection{\discintname}
The authors have no competing interests to declare that are relevant to the content of this article.
\end{credits}


\end{document}